# An Empirical Study of UMLS Concept Extraction from Clinical Notes using Boolean Combination Ensembles

Greg M. Silverman[1], Raymond L. Finzel[2], Michael V. Heinz[3], Jake Vasilakes[2,4], Jacob C. Solinsky[2], Reed McEwan[4], Benjamin C. Knoll[4], Christopher J. Tignanelli[1], Hongfang Liu[5], Hua Xu[6], Xiaoqian Jiang[6], Genevieve B. Melton[1,4], Serguei VS Pakhomov[2,4]

[1] Department of Surgery, University of Minnesota, Minneapolis, USA
[2] College of Pharmacy, University of Minnesota, Minneapolis, USA
[3] Geisel School of Medicine, Dartmouth College, Hanover, USA
[4] Institute for Health Informatics, University of Minnesota, Minneapolis, USA
[5] Department of Health Science Research, Mayo Clinic, Rochester, USA
[6] School of Biomedical Informatics, The University of Texas Health Science Center, Houston, USA

Corresponding Author:
Greg M Silverman
gms@umn.edu
Department of Surgery
MMC 195
420 Delaware St SE
Minneapolis, MN 55455

**Keywords**: Natural Language Processing, Information Extraction, Metathesaurus, Semantic Network, Algorithm

# Abstract

**Objective** To investigate behavior of Boolean operators on combining annotation output from multiple Natural Language Processing (NLP) systems across multiple corpora and to assess how filtering by aggregation of Unified Medical Language System (UMLS) Metathesaurus concepts affects system performance for Named Entity Recognition (NER) of UMLS concepts.

**Materials and methods** Three corpora annotated for UMLS concepts were used: 2010 i2b2 VA challenge set (31,161 annotations), Multi-source Integrated Platform for Answering Clinical Questions (MiPACQ) corpus (17,457 annotations including UMLS concept unique identifiers), and Fairview Health Services corpus (44,530 annotations). Our framework combines annotations generated by any number of NLP systems into an exhaustive set of ensembles using an approximate grid-search over combinations of Boolean operations. Performance of these Boolean combination ensembles was compared between all available named entity annotations to performance on annotation subsets filtered by UMLS semantic groups.

**Results** We demonstrated how optimized Boolean combination ensembles were constructed using the Fairview corpus on the collection of UMLS concepts filtered by the group *Procedures* and how our grid-search strategy identified 20 ensembles that outperformed all individual NLP systems for the group *Chemicals & Drugs*. We also showed that for UMLS concept matching, Boolean ensembling of the MiPACQ corpus trended towards higher performance over individual systems.

**Discussion** Boolean combination ensembles outperformed single systems in most cases. Use of an approximate grid-search can help optimize the precision-recall tradeoff and can provide a set of heuristics for choosing an optimal set of ensembles.

**Conclusion** Ensembling can improve NER performance over individual systems. The framework we developed can be used to tailor the choice of Boolean combination ensembles to a diverse set of tasks. Our results indicate that NER and concept mapping remain challenging problems for clinical NLP.

# INTRODUCTION

In this paper, we introduce NLP-Ensemble-Explorer, an evaluation framework that integrates output from multiple Natural Language Processing (NLP) systems and corpora as Boolean combination ensembles. NLP-Ensemble-Explorer utilizes a custom mapping to the Unified Medical Language System (UMLS) Semantic Network to categorize UMLS concepts by groups (e.g., findings, disorders, procedures, etc.)[1]. We investigated performance of these ensembles for Named Entity Recognition (NER) of UMLS concepts and matching of UMLS Concept Unique Identifiers (CUI)s.

Named entities (NE) can be mapped to UMLS CUIs and then extracted to provide consistent and unambiguous conceptual indexing for informatics research as discussed by Reátegui and Ratté, Pradhan *et al*., and Soldaini and Goharian, *et al*.[2][3][4]. Reátegui and Ratté compared MetaMap and cTAKES, showing the benefit of aggregation of concepts by UMLS semantic type[2], while Pradhan, et al. discuss the ShARe/CLEF eHealth 2013 Evaluation Lab Task 1, where NER and concept mapping on *Disorder* mention was done[3]. Soldaini *et al.* developed a fast UMLS lookup and extraction tool[4]. The Department of Veterans Affairs (VA) developed Sophia as a tool for fast UMLS concept extraction at large scale[5].

Ensembling of NLP systems leverages the complementary nature of statistical models and rule-bases underlying individual systems. If NLP systems are complementary, their ensembles should theoretically result in improved performance over any individual system. Indeed, multiple studies have shown improved results with ensembling for their target cohorts. In their study of ensembling methods for medical acronym disambiguation, Finley *et al.* observed increased ensemble accuracy (48.6%) compared to individual systems (42.1%)[6]. Silverman *et al.* and Tignanelli *et al*, showed that simple ensembling methods produced marked increases in precision and recall compared to individual systems for determining if appropriate treatment was provided in prehospital care [7][8]. Kuo *et al.* compared ensembling methods for extraction of UMLS concepts across different semantic groups, however, they found improvements in performance offered by ensembles are accompanied by high variability dependent on the target cohort, a qualification of ensemble performance that we hope to address in this current study[9].

More generally, Barreno et al. proved that in the space of optimized binary classifiers, Boolean AND (∧) and OR (∨) rules are always part of the optimal set of combined classifiers[10]. We developed NLP-Ensemble-Explorer to test this idea through examination of an exhaustive set of Boolean combinations of output from multiple NLP systems using logical ∧ and ∨ operators. To this end, we created ensembles by combining annotations from multiple clinical datasets using multiple out-of-the-box NLP systems. To the best of our knowledge, this study is the first to offer such a thorough evaluation of this type of ensembling. We demonstrate this strategy has potential to exploit strengths of each system and corpus and may provide a powerful tool for mapping unstructured text to controlled vocabularies like the UMLS.

# MATERIALS AND METHODS

## 2.1 Experimental task

We evaluated five widely available clinical NLP systems on the task of Named Entity Recognition (NER) of text spans representing UMLS concepts across different corpora. We also evaluated these systems on the task of concept matching of CUIs. Measures were computed for performance on individual system and Boolean combination ensembles of these systems. For NER, we also used majority voting ensembles. In addition, we analyzed performance of these systems on all annotations within each corpus as well as their subsets by semantic group (see McCray *et al*. for description of semantic groups[11]).

## 2.2 Corpora

We explored the idea of an "ensemble of corpora" to ensure we made as few assumptions about the stability of concepts (viz., concepts may drift over time[12] or may not have had consistent meaning between specialties); semantic categories (e.g., some concepts ambiguously belong to multiple categories[11]); and annotation tasks across gold standards (e.g., annotation guidelines may differ). Furthermore, Enríquez, *et al.*, showed that both corpus size and quality also affect performance of ensemble combinations of NLP systems[13]. Thus, while average performance across many related corpora is already a useful metric, performance variation between corpora in the ensemble could indicate particular NLP systems a) only perform well on particular note types, b) have a bias for annotations produced for a specific task, or c) semantic categories an NLP system purports to produce may encompass a slightly different semantic space than that of a particular gold standard.

For this study, two publicly available corpora were used: Multi-source Integrated Platform for Answering Clinical Questions (MiPACQ) and the 2010 i2b2 VA challenge set (referred to as "i2b2" hereafter). MiPACQ consists of fully anonymized clinical narratives from randomly selected Mayo Clinic clinical notes as well as fully anonymized Mayo Clinic pathology notes related to colon cancer that have been annotated with given annotation guidelines[14][15]. i2b2 consists of discharge summaries and progress notes taken from multiple independent institutions[16].

A third non-public corpus of manually annotated, fully identifiable outpatient clinical narratives from Fairview Health Services was also used. Notes in this corpus were randomly selected from the Fairview Epic Electronic Health Record for patients with congestive heart failure, type 2 diabetes mellitus, and chest pain. The entire patient medical record consisting of all notes for each patient up to 2010 was collected at the time of corpus creation and manually annotated for UMLS semantic types, acronyms, negation, and uncertainty as per guidelines[17] by two pharmacy students. An overlapping set of 15 patient records (37.5%) were annotated to calculate inter-rater reliability between the two raters (97% agreement, kappa 0.70).

*NB:* Some of the NLP systems included in this study are not completely independent of the annotations in MiPACQ and i2b2 corpora, since they used these corpora to train some of their modules[18][19]. None of the systems used any of the Fairview data for training. Therefore, we treat the evaluation results on MiPACQ and i2b2 as an upper limit on system performance and use these results only to compare ensembles with individual systems. The Fairview corpus was subsequently treated as an unbiased evaluation of systems and their ensembles

| Corpus/semantic type | # Annotations | # Notes |
|---|---:|---:|
| MiPACQ: All groups | 17,457 | 342 |
|     procedures | 2,861 | 307 |
|     disorder | 4,736 | 307 |
|     sign_symptom | 3,191 | 287 |
|     anatomy | 4,010 | 301 |
|     chemicals_and_drugs | 2,213 | 249 |
| I2b2: All groups | 31,161 | 256 |
|     test | 9,225 | 247 |
|     treatment | 9,334 | 254 |
|     problem | 12,592 | 255 |
| Fairview: All groups | 44,530 | 40 |
|     drug | 12,875 | 40 |
|     finding | 19,622 | 40 |
|     anatomy | 6,169 | 40 |
|     procedure | 1,643 | 40 |

**Table 1:** Summary of corpora used in this study

Table 1 provides a summary of the number of annotations, notes, and UMLS semantic types from each corpus (*NB*: *All groups* refers to the collection of annotations across all UMLS Semantic Types within a corpus). Annotations from each corpus were parsed and stored in a MySQL database for standardized formatting and ease of retrieval. Jupyter Notebooks for parsing each corpus are available[20].

## 2.3 NLP systems

Text notes from each corpus were processed using the NLP Artifact Discovery and Preparation Toolkit for Kubernetes (NLP-ADAPT-kube), which includes the following NLP systems compatible with the Unstructured Information Management Architecture (UIMA): The BioMedical Information Collection and Understanding System (BioMedICUS); the Clinical Language Annotation, Modeling, and Processing Toolkit (CLAMP); the Clinical Text Analysis and Knowledge Extraction System (cTAKES); and MetaMap (with UIMA adapter). Notes were also processed using QuickUMLS, a non-UIMA NLP system for concept extraction[21][22][23][24][25][26][27].

All text notes were first processed using a script to convert all text to plain text (ASCII) encoding as required by MetaMap[28]. For this study, we utilized standard pipelines available in all systems for annotation of UMLS concepts. To minimize false positives, as determined by our prior experience with these systems, we used 800 as the threshold for MetaMap's evaluation score[29]. For QuickUMLS we used 0.80 as the threshold for the Jaccard similarity index and the default of highest score for selecting best matches in the case of overlapping concepts. For the five systems used in this study, only MetaMap had word sense disambiguation available for filtering out ambiguous mappings to the UMLS[30].

Once annotations were generated for each system, multiple overlapping concepts were further disambiguated in a custom method available as part of the NLP-Ensemble-Explorer framework. This method used the following hierarchical rules to determine "best" match: Choose longest overlapping span if it exists; else for systems with probability scoring for likelihood of given UMLS concept use highest score (NB: the only system with no likelihood scoring was cTAKES); else randomly shuffle multiple concepts mapped to span and select concept in first position[31].

The pipeline for NLP-Ensemble-Explorer used the library dkpro-cassis[32], developed by the Technische Universität Darmstadt, to convert annotation objects produced by NLP-ADAPT-kube in the UIMA XML Metadata Interchange Common Analysis System format to standard python objects for post processing. After all annotation objects were processed they were pooled together and stored in a MySQL database for integration into the NLP-Ensemble-Explorer pipeline..

## 2.4 Semantic type mapping

To assess performance of NLP systems, output of the systems first had to be constrained to categories of semantic types present in the gold standards. Where systems provided their own notion of "semantic type" and where those categories mapped reasonably well to the categories in the gold standards, output was constrained using these system-specific categories. In cases where the systems provided UMLS Semantic Types but no bespoke semantic categories, we relied on Semantic Type Mappings and Semantic Groups[33] developed at the National Library of Medicine to subdivide annotations by semantic type. This strategy, as illustrated in Figure 1, enabled a comprehensive hierarchical mapping of semantic types to semantic groups to allow the combination and comparison of multiple corpora and systems.

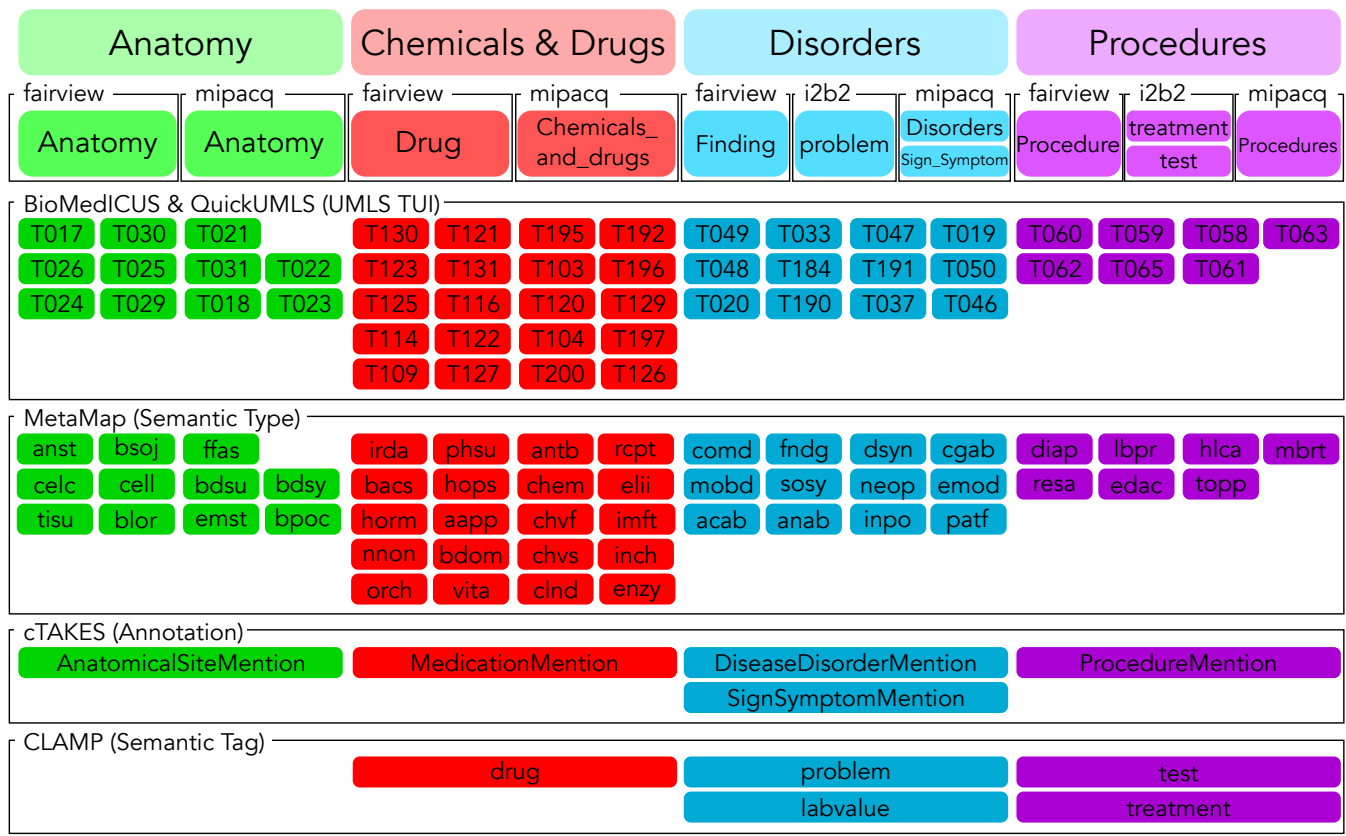

**Figure 1**

At the top level are the UMLS semantic groups that aligned best with the semantic types present in each gold standard, which appear at the second level in Figure 1. The remainder of Figure 1 illustrates semantic types available in individual NLP systems and how they were mapped to the top-level semantic groups.

## 2.5 UMLS versions

The following versions of the UMLS, by system, were used: 2019AB by QuickUMLS and MetaMap; 2016AB by cTAKES; 2016AA by BioMedICUS; 2014AB by CLAMP.

## 2.6 NLP-Ensemble-Explorer

NLP-Ensemble-Explorer is available as a Jupyter notebook[20] and was designed as part of an analytics pipeline for evaluating ensembling performance for NER and IE, but is extensible to other NLP tasks. This framework is based in part on The Amicus Metasystem for Interoperation and Combination of UIMA Systems (AMICUS) developed by Finley[34] and the Natural Language Processing Type and Annotation Browser (NLPTAB) developed by Knoll[35]. AMICUS and NLP-TAB were developed to be used with NLP-ADAPT for evaluating and combining annotation output between UIMA-based NLP systems. These systems were intended to examine small subsets of annotations from a larger corpus for selection of optimal annotation types for use in larger scale tasks. NLP-Ensemble-Explorer was developed independent of system architecture as a successor to these tools. For more information on the expected format of annotation data for use by NLP-Ensemble-Explorer see the README.md[20].

### 2.6.1 Boolean combination and majority vote ensembles

Comprehensive lists of all permutations for NLP systems were transformed into an exhaustive set of Boolean combinations using the logical ∨ operator - to represent a UNION set operation (or ∪); and the logical ∧ operator - to represent an INTERSECTION set operation (or ∩). Equivalence of logical and set theoretic operators is given by logical conjunction and disjunction of sets[36][37]. Boolean combination ensembles were then evaluated to produce a merged set of system annotations to assess performance against gold standard annotations. Once a Boolean expression was generated it was stored and evaluated as a binary tree using the parse tree algorithms provided by Miller and Ranum[38]. To evaluate majority vote ensembles we used labels on which the majority of the NLP systems agreed. In the case of a tie, the label was chosen at random.

Please see Appendix A for a description of our approximate grid-search method for determining optimal Boolean combination ensembles and results.

### 2.6.2 Cross-system semantic group union merge

This task involved evaluation of an ensemble using the ∨ operation between different semantic groups annotated by different individual NLP systems within a corpus. This purpose of this task was to assess variation against single system performance for the *All groups* category and to test it as a possible vehicle for creation of more complex Boolean combinations.

### 2.6.3 UMLS CUI matching

We also evaluated the tasks of document and mention-level CUI matching within the MiPACQ corpus using all systems, including biased (BioMedICUS and cTAKES), and unbiased only (CLAMP, MetaMap and QuickUMLS) to see if ensembling offered improvement subsequent to NER. Our objective here was not to compare individual systems but to control for as much variability as possible in determining whether ensembles would outperform individual systems.

### 2.6.4 Matching algorithms

In a recent 2018 n2c2 challenge that involved NER tasks, the preferred strategy for identifying matches between NLP system annotations and gold standard annotations relied on relaxing the boundaries of matches to give a full match[39]. We used a character-level binary i-o classification partial matching scheme (labeled as 1 and 0, respectively) to adjust the weight based on the length of the match to appropriately weight matches to the number of characters in overlap between an annotated span in the system and gold standard span set for each document[40].

For document-level CUI extraction, we used a scheme similar to Kuo, et al.[9] where a match is counted if for the list of CUIs in a corpus a specific CUI appears in both the set of system and gold standard annotations within the same document, independent of an annotation. However, while they treated this as a binary classification task, we used a multilabel multiclass classifier to account for multiple labels within the same document. For mention-level CUI extraction, we used a multiclass classifier with the i-o labelling scheme described above. In the case where multiple CUIs are assigned to the same or overlapping text span, when doing a UNION operation, we used a majority voting scheme with the longer text span serving as a tiebreaker followed by a random selection in case of equal span lengths but different CUIs within each system,

### 2.6.5 Statistical analyses

Standard performance measures of precision, recall, and their harmonic mean (F1-score) were used to evaluate individual systems and their Boolean combination ensembles. We used 95% Bernoulli confidence intervals for NER as a test for statistical significance as proposed by Witten and Frank[41][42]. Following this methodology, when there is overlap between confidence intervals, we do not reject the *null* hypothesis of there being no difference between measures.

### 2.6.6 Complementarity

Complementarity, as outlined by Derczynski, has the following characteristics: a) High disagreement is a strong indicator of error resulting from how different approaches to many tasks that use similar information achieve similar performance; b) Complementarity is additive, such that as more systems are added and the performance of the hypothetically best possible combination increases, the complementary rate of the next system decreases monotonically; and c) As the amount of information common to all systems increases (e.g. through training data), complementarity decreases[43].

The complementary rate. as originally defined by Brill and Wu, and elaborated on by Derczynski, was used as a measure of the maximum potential improvement a system has in the NER task over another system[43,44]. Furthermore, as discussed by Derczynski, modified measures of precision, recall and F1-score were used to quantify the effect complementarity of individual systems have on each other in terms of one system detecting items that the other did not [43]. See equations 1-4, respectively.

# RESULTS

We assessed performance of individual systems and Boolean combination ensembles on the tasks of concept span identification, and CUI matching at both document and mention-level for each corpus as a whole and by semantic groups (see Figure 1). For the NER task, we also used a majority vote ensemble for concept span identification.

## 3.1 Individual system performance

|  |  | SYSTEM | | | | | | | | | | | | | |
|---|---|---|---|---|---|---|---|---|---|---|---|---|---|---|---|
| Corpus | | BioMedICUS | | | CLAMP | | | cTAKES | | | MetaMap | | | QuickUMLS | | |
| Semantic Group | n | p | r | F1 | p | r | F1 | p | r | F1 | p | r | F1 | p | r | F1 |
| **Fairview** | | | | | | | | | | | | | | | | |
| Anatomy | 47131 | 0.21 | 0.69 | 0.33 | n/a | n/a | n/a | **0.3** | **0.77** | **0.44** | 0.27 | 0.43 | 0.34 | 0.23 | 0.48 | 0.22 |
| Chemicals & Drugs | 121969 | 0.36 | 0.64 | 0.46 | **0.48** | 0.79 | **0.61** | 0.26 | **0.89** | 0.41 | 0.21 | 0.48 | 0.30 | 0.38 | 0.81 | 0.52 |
| Disorders | 325189 | 0.3 | 0.52 | 0.38 | **0.45** | 0.57 | **0.50** | 0.37 | **0.55** | 0.45 | 0.32 | 0.44 | 0.37 | 0.33 | 0.47 | 0.39 |
| Procedures | 22747 | 0.08 | **0.57** | 0.14 | 0.05 | 0.54 | 0.09 | **0.1** | 0.53 | **0.17** | 0.05 | 0.5 | 0.09 | 0.05 | 0.55 | 0.10 |
| All groups | 469619 | 0.27 | 0.64 | 0.38 | 0.32 | 0.68 | 0.43 | **0.33** | **0.72** | **0.45** | 0.14 | 0.59 | 0.23 | 0.21 | 0.68 | 0.32 |
| **i2b2** | | | | | | | | | | | | | | | | |
| Disorders | 213599 | 0.54 | 0.53 | 0.54 | **0.91** | **0.98** | **0.95** | 0.62 | 0.53 | 0.58 | 0.42 | 0.51 | 0.47 | 0.57 | 0.61 | 0.59 |
| Procedures | 223708 | 0.76 | 0.32 | 0.45 | **0.98** | **0.63** | **0.77** | 0.78 | 0.25 | 0.38 | 0.41 | 0.31 | 0.36 | 0.58 | 0.33 | 0.42 |
| All groups | 437307 | 0.63 | 0.63 | 0.63 | **0.83** | **0.9** | **0.87** | 0.73 | 0.61 | 0.67 | 0.41 | 0.63 | 0.50 | 0.59 | 0.7 | 0.64 |
| **MiPACQ** | | | | | | | | | | | | | | | | |
| Anatomy | 25166 | 0.71 | 0.78 | 0.75 | n/a | n/a | n/a | **0.83** | **0.86** | **0.85** | 0.71 | 0.46 | 0.57 | 0.69 | 0.78 | 0.74 |
| Chemicals & Drugs | 18242 | 0.67 | 0.74 | 0.72 | **0.72** | 0.66 | 0.70 | 0.62 | **0.91** | **0.75** | 0.47 | 0.64 | 0.55 | 0.66 | 0.77 | 0.72 |
| Disorders | 74704 | 0.59 | 0.7 | 0.65 | 0.56 | **0.82** | 0.67 | **0.66** | 0.75 | **0.71** | 0.56 | 0.7 | 0.63 | 0.57 | 0.79 | 0.67 |
| Procedures | 31705 | 0.57 | 0.6 | 0.59 | 0.41 | 0.65 | 0.51 | **0.67** | 0.57 | 0.62 | 0.47 | 0.64 | 0.55 | 0.51 | **0.72** | **0.61** |
| All groups | 138755 | 0.62 | 0.77 | 0.69 | 0.57 | 0.78 | 0.67 | **0.72** | 0.78 | **0.75** | 0.39 | 0.79 | 0.53 | 0.56 | **0.84** | 0.68 |

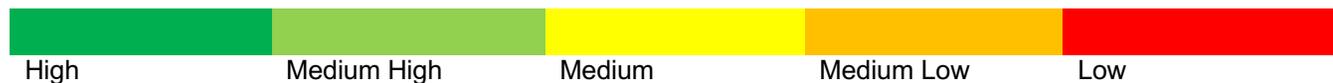

| High | Medium High | Medium | Medium Low | Low |

**Table 2:** Individual system performance for NER by corpus and semantic group; **n** refers to total number gold standard characters as used in calculation of measures: **p**=precision, **r**=recall, **F1**=F1-score; Best row value for metric is bolded; row values of F1-score color coded for High to Low as per legend.

Confidence intervals for each individual system's F1-score can be found in Appendix B, Table 1.

Differences in performance between the whole corpus (*All Groups)* and lowest performing individual semantic categories was often marked, with differences around 0.30 common among F1-Scores. This was especially true for the previously unused Fairview corpus. Moreover, individual NLP systems each had distinct strengths and weaknesses, with top performers for one semantic group often struggling in other areas (e.g., CLAMP for Fairview *Chemicals & Drugs* and *Procedures*, respectively).

Meanwhile, performance of individual systems on corpora that had previously served as training sets for those same systems was often more consistent. This indicates there is complementarity between individual NLP systems that could be harnessed by a variety of ensemble methods, which may be especially pronounced for novel corpora and new domains.

Results presented below represent a selection of these methods in action for their effect on F1-score for the given task (unless otherwise specified).

### 3.1.1 Complementarity

## 3.2 Cross-system semantic group union merge for Fairview corpus

This method inherently utilizes complementarity between systems that perform best on a particular semantic group. The highest F1-score was achieved by CLAMP to annotate *Chemicals & Drugs* and *Disorders* and cTAKES to annotate for *Anatomy,* and creating a UNION merge of the two sets of annotations (with precision 0.45, recall 0.69, and F1-score 0.54), The results were statistically significant in comparison to *All groups* for the single systems CLAMP and cTAKES (an improvement of 0.09 over the next highest F1-score: F1: 0.541 in CI: (0.539, 0.543))

## 3.3 Boolean combinations for NER

We evaluated combinations of all 5 systems on all corpora and their semantic groups with the exceptions for *Anatomy*, where only 4 systems were evaluated (CLAMP does not annotate for *Anatomy*). For systems trained on a corpus (biased), viz., CLAMP for i2b2, and BioMedICUS and cTAKES for MiPACQ, all systems minus those noted were used in the unbiased evaluation. The total number of potential Boolean combinations when combining 5 systems was 2,425; for 4 systems it was 238; for 3 systems it was 28; and for 2 systems it was 4[45]. The results are shown in Table 3.

| Corpus | Highest F1-score | | | | Highest precision | | | | Highest recall | | | |
|---|---|---|---|---|---|---|---|---|---|---|---|---|
| Semantic Group | combination | p | r | F1 | combination | p | r | F1 | combination | p | r | F1 |
| **Fairview** | | | | | | | | | | | | |
| Anatomy | (((A∧E)∨D)∧C) | 0.48 | 0.47 | 0.48 | ((C∧D)∧E) | 0.64 | 0.20 | 0.30 | (((A∨C)∨D)∨E) | 0.20 | 0.84 | 0.32 |
| Chemicals & Drugs | ((A∨C)∧B) | 0.52 | 0.72 | 0.61 | ((A∧B)∧D) | 0.72 | 0.16 | 0.26 | ((((A∨B)∨C)∨D)∨E) | 0.19 | 0.95 | 0.32 |
| Disorders | ((((A∧C)∧D)∧E)∨B) | 0.44 | 0.60 | 0.51 | (((B∧C)∧D)∧E) | 0.66 | 0.09 | 0.16 | ((((A∨B)∨C)∨D)∨E) | 0.24 | 0.75 | 0.36 |
| Procedures | ((((B∨D)∨E)∧A)∧C) | 0.15 | 0.41 | 0.22 | (((A∧B)∧C)∧E) | 0.19 | 0.13 | 0.15 | ((((A∨B)∨C)∨D)∨E) | 0.03 | 0.79 | 0.06 |
| All groups | (((A∧D)∨B)∧C) | 0.43 | 0.54 | 0.48 | (((A∧B)∧C)∧D) | 0.56 | 0.18 | 0.28 | ((((A∨B)∨C)∨D)∨E) | 0.14 | 0.91 | 0.25 |
| **i2b2 *** | | | | | | | | | | | | |
| Disorders | ((((A∧C)∧D)∧E)∨B) | 0.85 | 0.98 | 0.91 | ((B∧C)∧E) | 1.00 | 0.27 | 0.42 | ((((A∨B)∨C)∨D)∨E) | 0.47 | 0.98 | 0.64 |
| Procedures | ((((A∧C)∧D)∧E)∨B) | 0.95 | 0.64 | 0.76 | ((B∧D)∧E) | 0.99 | 0.14 | 0.25 | ((((A∨B)∨C)∨D)∨E) | 0.54 | 0.72 | 0.62 |
| All groups | ((((A∧C)∧D)∧E)∨B) | 0.78 | 0.91 | 0.84 | (((A∧B)∧C)∧E) | 0.97 | 0.32 | 0.48 | ((((A∨B)∨C)∨D)∨E) | 0.46 | 0.95 | 0.62 |
| **MiPACQ *** | | | | | | | | | | | | |
| Anatomy | (((A∧D)∧E)∨C) | 0.80 | 0.87 | 0.83 | (((A∧C)∧D)∧E) | 0.93 | 0.29 | 0.44 | (((A∨C)∨D)∨E) | 0.65 | 0.92 | 0.76 |
| Chemicals & Drugs | ((((A∧D)∨B)∨E)∧C) | 0.73 | 0.78 | 0.76 | (((A∧B)∧C)∧E) | 0.82 | 0.58 | 0.68 | ((((A∨B)∨C)∨D)∨E) | 0.45 | 0.94 | 0.61 |
| Disorders | ((B∧D)∨(C∧E)) | 0.74 | 0.74 | 0.74 | ((((A∨D)∧B)∧C)∧E) | 0.91 | 0.34 | 0.50 | ((((A∨B)∨C)∨D)∨E) | 0.40 | 0.91 | 0.55 |
| Procedures | ((B∨E)∧(C∨D)) | 0.59 | 0.69 | 0.64 | (((B∧C)∧D)∧E) | 0.78 | 0.23 | 0.36 | ((((A∨B)∨C)∨D)∨E) | 0.34 | 0.86 | 0.49 |
| All groups | ((((A∨D)∧B)∧E)∨C) | 0.70 | 0.81 | 0.76 | (((A∧B)∧C)∧E) | 0.86 | 0.37 | 0.52 | ((((A∨B)∨C)∨D)∨E) | 0.37 | 0.95 | 0.53 |
| **i2b2 **** | | | | | | | | | | | | |
| Disorders | ((C∧D)∨E) | 0.56 | 0.62 | 0.59 | (((A∧C)∧D)∧E) | 0.77 | 0.28 | 0.41 | (((A∨C)∨D)∨E) | 0.40 | 0.69 | 0.51 |
| Procedures | ((A∨C)∨E) | 0.60 | 0.40 | 0.49 | (((A∧C)∧D)∧E) | 0.88 | 0.13 | 0.22 | (((A∨C)∨D)∨E) | 0.44 | 0.46 | 0.45 |
| All groups | ((A∧E)∨C) | 0.67 | 0.68 | 0.67 | ((A∧C)∧E) | 0.79 | 0.50 | 0.61 | (((A∨C)∨D)∨E) | 0.44 | 0.82 | 0.58 |
| **MiPACQ **** | | | | | | | | | | | | |
| Anatomy | (D∨E) | 0.66 | 0.85 | 0.75 | (D∧E) | 0.83 | 0.35 | 0.50 | (D∨E) | 0.66 | 0.85 | 0.74 |
| Chemicals & Drugs | ((B∨D)∧E) | 0.74 | 0.70 | 0.73 | (B∧E) | 0.78 | 0.61 | 0.68 | ((B∨D)∨E) | 0.48 | 0. | 0.62 |
| Disorders | ((B∨D)∧E) | 0.68 | 0.67 | 0.68 | ((B∧D)∧E) | 0.86 | 0.33 | 0.48 | ((B∨D)∨E) | 0.42 | 0.89 | 0.57 |
| Procedures | ((B∧D)∨E) | 0.51 | 0.75 | 0.61 | ((B∧D)∧E) | 0.71 | 0.31 | 0.43 | ((B∨D)∨E) | 0.35 | 0.85 | 0.50 |
| All groups | ((B∧D)∨E) | 0.55 | 0.86 | 0.68 | ((B∧D)∧E) | 0.80 | 0.36 | 0.50 | ((B∨D)∨E) | 0.37 | 0.95 | 0.53 |

**Table 3**: Boolean combination ensemble performance by corpus and semantic group for NER; Systems A-E, respectively are A. BioMedICUS, B. CLAMP, C. cTAKES, D. MetaMap and E. QuickUMLS; measures are **p**=precision, **r**=recall, **F1**=F1-score. * denotes evaluation with biased systems; ** denotes evaluation with unbiased systems.

Confidence intervals for each Boolean combination ensemble by highest given metric can be found in Appendix B, Table 2.

The highest performing Boolean combination ensembles exploit complementarity of individual systems through appropriate mixtures of the logical ∧ and ∨ operators. The following examples are representative of well-known patterns that boost performance through: a) Increasing precision by filtering out false positives through use of the ∧ operator; or b) Increasing recall by filtering out false negatives through use of the ∨ operator. Optimized Boolean combinations for higher F1-scores employ a balanced mix of these patterns. These examples also illustrate how complementarity can be beneficial when applied to Boolean combination ensembling. Unless otherwise noted, all differences given are statistically significant.

The Boolean combination ensemble ((((clamp∨metamap)∨quick_umls)∧biomedicus)∧ctakes) on the Fairview corpus for the semantic group *Procedures* provided the largest gain (0.08) over any single system (cTAKES). The *All groups* Boolean combination ensemble (((biomedicus∧metamap)∨clamp)∧ctakes) for the same corpus provided a 0.03 absolute improvement over any single system, and this ensemble bore a similar structure to the ensemble above, wherein an INTERSECTION operation with the cTAKES annotation set attenuated the results from several other systems. Further assessment of the individual systems is warranted in order to understand the ways in which these systems can be complementary, e.g. understanding what work QuickUMLS is doing in the ensemble that improves performance for *Procedures* that is no longer helpful when assessing performance on the corpus as a whole.

Boolean combination ensembles for MiPACQ with systems trained on this corpus were all either not significant or lower for F1-score compared to individual systems. Thus, these systems represent an upper bound. However, optimization on precision and recall is still possible, e.g., the Boolean combination (((biomedicus∧ctakes)∧metamap)∧quick_umls) for *Anatomy* gives a boost of 10% on precision. Similarly, for I2b2 we find the same consistent behavior, e.g., the combination ((clamp∧ctakes)∧quick_umls) (which is essentially a unanimous vote of the three systems) for *Disorders* boosted precision to 1.0.

## 3.4 Majority vote

The results of evaluating majority vote ensembles of NLP systems are shown in Table 4. We evaluated voting combinations for all 5 systems on all corpora and their semantic groups with the exceptions for *Anatomy*, where only 4 systems were evaluated (CLAMP does not annotate for *Anatomy*).

| Corpus | | | | |
|---|---|---|---|---|
| Semantic Group | systems | p | r | F1 |
| **Fairview** | | | | |
| Anatomy | A,C,D,E | 0.51 | 0.34 | 0.41 |
| Chemicals & Drugs | A,B,C,D,E | 0.83 | 0.45 | 0.58 |
| Disorders | A,B,C,D,E | 0.53 | 0.43 | 0.48 |
| Procedures | A,B,C,D,E | 0.58 | 0.10 | 0.18 |
| All groups | A,B,C,D,E | 0.72 | 0.32 | 0.44 |
| **i2b2** | | | | |
| Disorders | A,B,C,D,E | 0.59 | 0.66 | 0.62 |
| Procedures | A,B,C,D,E | 0.32 | 0.82 | 0.46 |
| All groups | A,B,C,D,E | 0.71 | 0.70 | 0.70 |
| **MiPACQ** | | | | |
| Anatomy | A,C,D,E | 0.76 | 0.82 | 0.79 |
| Chemicals & Drugs | A,B,C,D,E | 0.78 | 0.72 | 0.75 |
| Disorders | A,B,C,D,E | 0.78 | 0.67 | 0.72 |
| Procedures | A,B,C,D,E | 0.65 | 0.63 | 0.64 |
| All groups | A,B,C,D,E | 0.84 | 0.65 | 0.73 |

**Table 4:** Majority vote ensemble performance by corpus and semantic group for NER; Systems A-E, respectively are A. BioMedICUS, B. CLAMP, C. cTAKES, D. MetaMap and E. QuickUMLS; measures are **p**=precision, **r**=recall, **F1**=F1-score.

## 3.5 UMLS CUI matching

We used systems where the MiPACQ corpus was not part of the development pipeline for this analysis (CLAMP, MetaMap and QuickUMLS). The total number of potential Boolean combinations when combining 3 systems was 28, while for 2 systems there was 4[45]. Results for Boolean combination ensembles and single system performance are shown in Table 5.

| Corpus | Highest F1-score | | | | Highest F1-score | | | |
|---|---|---|---|---|---|---|---|---|
| Semantic Group | combination | p | r | F1 | system | p | r | F1 |
| MiPACQ * | | | | | | | | |
|   Anatomy | (DVE) | 0.45 | 0.51 | 0.46 | E | 0.37 | 0.43 | 0.38 |
|   Chemicals & Drugs | ((BVD)VE) | 0.71 | 0.79 | 0.72 | E | 0.62 | 0.70 | 0.64 |
|   Disorders | ((BVD)VE) | 0.58 | 0.64 | 0.58 | E | 0.52 | 0.55 | 0.51 |
|   Procedures | ((BVD)VE) | 0.50 | 0.59 | 0.52 | E | 0.45 | 0.50 | 0.46 |
|   All groups | ((BVD)VE) | 0.57 | 0.65 | 0.58 | E | 0.50 | 0.55 | 0.50 |
| MiPACQ ** | | | | | | | | |
|   Anatomy | (DVE) | 0.39 | 0.47 | 0.39 | E | 0.33 | 0.40 | 0.33 |
|   Chemicals & Drugs | ((BVD)VE) | 0.68 | 0.76 | 0.69 | E | 0.61 | 0.68 | 0.62 |
|   Disorders | ((BVD)VE) | 0.52 | 0.56 | 0.51 | E | 0.47 | 0.49 | 0.45 |
|   Procedures | ((BVD)VE) | 0.46 | 0.52 | 0.46 | E | 0.41 | 0.45 | 0.41 |
|   All groups | ((BVD)VE) | 0.55 | 0.63 | 0.55 | E | 0.47 | 0.54 | 0.48 |

**Table 5**: Comparison of best performing unbiased Boolean combinations and single systems by corpus and semantic group for CUI matching; * denotes comparison for document-level CUI matching; ** denotes comparison for mention-level CUI matching; Unbiased systems, respectively are B. CLAMP, D. MetaMap, and E. QuickUMLS; measures are **p**=precision, **r**=recall, **F1**=Macro F1-score.

The magnitude of F1-score for Boolean combination ensembles, for both corpus and mention-level CUI matching, appears to be greater than for single systems. While no significance testing was done, the differences for most comparisons within ensembles and single systems (comparing *All groups* to each semantic group for both tasks) were noteworthy, with *Chemicals & Drugs* alone higher than *All groups*. Comparisons between single systems and Boolean combination ensembles appear to trend in favor of ensembles, as ensembles demonstrate higher performance on nearly all measures.

# DISCUSSION

This study used an extensible framework to create and evaluate a particular class of NLP system ensembles for the tasks of NER and matching labeled NEs. We investigated an empirical method in which examination of all possible Boolean combinations between NLP systems used an approximate grid-search to determine the cluster of optimal Boolean combinations. The results of this study show this strategy has potential to produce ensembles that outperform individual systems on all measures.

Our results exemplify the complementary strengths and weaknesses of individual systems across corpora. For the Fairview corpus a major weakness was that all individual systems (with the exception of MetaMap) had lower performance on *Anatomy* in comparison to *All groups*. Thus, it was no surprise that the highest performing Boolean combination ensemble was not statistically different than *All groups*, and that the most significant improvement was achieved by other means (the cross-system semantic group union merge). In contrast, complementarity of these same systems was revealed by the higher performance on *Anatomy* for the MiPACQ corpus. For Fairview *Disorders,* the results were mixed. BioMedICUS and cTAKES did not offer any improvement over *All groups* (echoing similar results for MiPACQ), though others did. Lastly, for *Chemicals & Drugs* the only system that had lower performance than *All groups* for both Fairview and MiPACQ was cTAKES.

Our results for QuickUMLS, cTAKES and MetaMap were consistent with previous work on the i2b2 corpus by Soldaini and Goharian [4]. Our results for MetaMap were similar for MetaMap and slightly higher for cTAKES than for Divita, *et al.*, on the i2b2 corpus for *Disorders*[5]*.* Our results for MetaMap were consistent with those reported by Kim and Rilof, who observed a significant difference between i2b2 annotations and MetaMap's concept boundary definition[46].

## 4.1 Boolean combination ensembles for NER

For cases where the number of combined systems was less than 3, union operations significantly increased recall while decreasing precision with the exception of Fairview *Procedures*. These results aligned to the work of Silverman, *et al.*[7]*,* Finley, *et al.*[6], and Kuo, *et al.*, where there was a consistent boost in recall for union merge for the specified tasks. Our experience of a precision-recall tradeoff when using intersection ensembles aligned with Kuo, *et al.[9].* However, in the case where 3 or more systems are combined, especially when limiting the analysis to a particular semantic group (e.g., Fairview *Chemicals & Drugs*), the imbalance between precision and recall diminishes. Unlike other studies, including that of Kang, *et al.* and Liu *et al.*, that showed use of simple intersection merges diminished concept extraction performance, especially in comparison to a simple majority voting scheme[47][48], our study has shown use of intersection with union operations to be very beneficial.

Indeed, Boolean combination ensembles have potential to boost performance through complementarity as demonstrated by comparing ensembles with "Highest F1-score" to individual systems for semantic groups of the Fairview corpus.

Individual systems often had outsized influence on performance, e.g., the Boolean combination ((biomedicus∨ctakes)∧clamp) for Fairview *Chemicals & Drugs* was comprised of CLAMP, which individually had the highest F1-score and precision for this category, while cTAKES had highest recall, and thus when combined with a union of BioMedICUS a higher recall helped boost F1-score. This seems reasonable since the combination is interpretable as "if text is found both in the CLAMP annotations and in either the BioMedICUS or cTAKES annotations." An informal examination of our data found a similar trend of mixed operations for other Boolean combinations, including the ensemble combination ((((biomedicus∧ctakes)∧metamap)∧quick_umls)∨clamp), which had the highest F1-score for Fairview *Disorders* (under column denoted "Highest F1") and all biased i2b2 ensembles and their semantic groups. Boolean combinations having a mix of intersections and unions were often the most effective, as predicted by the theoretical framework of Barreno, *et al.*[10].

## 4.2 Majority vote ensemble

The combination of NLP systems via the majority vote approach was found to produce less consistent improvement compared to simple Boolean combinations, which was consistent with those reported by Finley, *et al.*[6], and those of Shahzad and Lavesson, in which they compare performance of various voting ensemble methods for use in malware detection[49]. Our results also show that even in cases where ensembles of NLP systems do not improve the overall performance over individual systems, there may still be an improvement for specific semantic groups of named entities (e.g., Chemicals & Drugs).

## 4.3 UMLS CUI matching

Results for the task of UMLS CUI matching were encouraging. Generally, ensembles for both document/mention-level CUI matching trended towards higher performance over single systems. It should be noted, that for both document/mention-level CUI matching, the only ensemble that appears to have performed better than *All groups* was *Chemicals & Drugs*.

Examination of document-level matching indicates all systems performed better on *Chemicals & Drugs* compared to *All groups*. Similar results were obtained for other groups, with the exception of CLAMP performing worse on *Procedures*. For mention-level matching, all systems, except for CLAMP, performed better than *All groups* (results for other groups were more varied).

Also of note, the single system that appeared to have the highest F1-score was QuickUMLS, while the same ensembles for each semantic group appeared to have the highest F1-score. For the semantic group *Anatomy* at both the document/mention-level this was (metamap∨quickumls); while for all other semantic groups this was ((clamp∨metamap)∨quickumls). Whether these ensembles were of the same form was due to characteristics of the corpus or the particular task or the number of systems used remains an open question.

These results are consistent with the findings of Kuo, *et al.*[9] for their basic ensemble, both at the document/mention-level for extraction tasks. Our results for *Chemicals & Drugs* also align with those of Reátegui and Ratté where use of aggregation by semantic group has potential to increase performance of extraction of CUIs[2]. Furthermore, for the task of *Disorder* mapping, our results indicate that this task was congruent with the results of Pradhan, *et al.*, in which concept mapping was much more difficult for this semantic group than was span identification[3].

## 4.4 Sources of complementarity

Likely sources of complementarity between individual NLP systems can be found in the domain each system was designed to work in, including the types of notes each system was trained on, and the way that each system does UMLS lookup.

Each system used in the study was designed for use in a specific domain. BioMedICUS, CLAMP and cTAKES were each designed to work with clinical notes[23][24][25]. MetaMap was designed originally for indexing of biomedical literature but has since added features allowing it to work with clinical text[50][51][52]. QuickUMLS, on the other hand, was designed primarily as a lookup tool independent of any particular clinical domain[4].

Furthermore, as previously noted, both cTAKES and BioMediCUS were trained on the MiPACQ corpus, which comprised of randomly selected clinical notes and pathology notes for colon cancer[14][15], and CLAMP was trained on the i2b2 corpus, which were discharge summaries culled from multiple institutions[16].

For details on how each system does UMLS concept matching and lookup, please see Appendix C.

## 4.5 Limitations

This work has some limitations. First, larger ensembles and more complex tasks may be computationally challenging necessitating further optimizations. Second, our evaluation was limited to NER and concept matching tasks and did not include other related tasks such as identification of concept modifications (e.g., negation, certainty, status, etc.). However, preliminary experiments in our lab using Boolean combination ensembles for use in detecting UMLS concept negation has shown potential to increase accuracy. Third, our choice of semantic mappings relied on existing categories from individual NLP systems and from the UMLS Semantic Network. Other mappings that better represent concepts found in the gold standards may exist. Fourth, due to use of built-in python methods, which did not allow for ease of extraction of required parameters, we did not perform significance tests for CUI matching tasks. However, our results can still be used to observe trends. Fifth, since the corpus we used for concept matching of CUIs was used for training statistical models for two of the NLP systems used in this study, it was not possible to perform an unbiased comparison of Boolean combination ensembles with more than three systems. Thus, use of a gold standard corpus labeled for CUIs independent of any NLP system training should be used for a more rigorous evaluation of larger Boolean combination ensembles and their behavior. Sixth, use of multiple versions of the UMLS is a potential confounding factor in this study. However, we observed that performance of systems configured with same versions of UMLS varied greatly across corpora and semantic groups. In addition, systems using older versions of UMLS performed much better than those with newer versions of UMLS on some tasks. It is likely that different UMLS versions contribute to complementarity, since, for example, developers of SemRep continue to use 2006AA, as opposed to newer versions of the UMLS, because there are fewer concepts/synonyms which decreases ambiguity[53]. Lastly, our test for complementarity was only used for the NER task… it is important to note that the goal of this study was not to focus on individual system performance, but instead to examine ensemble performance trends for different tasks across corpora and semantic groups using default pipelines of NLP systems.

## 4.6 Applicability of results

Our observations indicate Boolean combination ensembles for NER will produce a mix of ∧ and ∨ operators for top performing ensembles for F1-score, whereas for CUI matching, best ensembles are more likely to have systems joined with the same operator, at least when using only 3 systems. NER generally outperforms document-level matching of CUIs which subsequently outperforms mention-level matching of CUIs. However, for both document and annotation-level CUI matching, most differences between individual systems and the ensembles with highest F1-score appeared to be greater than those for NER.  For NER, most top performing Boolean combinations performed better than individual systems for the same semantic group; for both document/mention-level matching of CUIs, all top performing Boolean combination ensembles appeared to outperform single systems across all semantic groups.

Methods used in this study can be directly applied to any other clinical NLP systems that map text to UMLS concepts (or other vocabularies) and will work for evaluating any corpus annotated for UMLS concepts (or other vocabularies). While results in this paper are applicable to particular annotation tasks, NLP systems and corpora, the NLP-Ensemble-Explorer framework is designed to be extensible for other clinical annotation tasks. For example, our lab has successfully used NLP-Ensemble-Explorer for the task of clinical acronym sense disambiguation on a completely different corpus [link to AMIA workshop and github branch and corpus].

It should be noted that this study applied three distinct corpora, with each consisting of different types of clinical notes with distinct patient cohorts. As noted by Kuo, *et al.*[10] target cohort adds variability to ensembles. We believe these targeted differences could be beneficially harnessed within a private set of notes from different departments within a single institution with focus on specific semantic groups.

Furthermore, methods employed in this study can be utilized as heuristics for evaluating independent annotated corpora. See Appendix C for a suggested protocol on how to apply this to a desired NER/IE task.

### 4.7 Future work

Equivalence of majority-vote to Boolean combinations?

While results of our study determined the optimal NLP system or Boolean combination ensemble of NLP systems for a particular task, it is necessary to show our results align with those of Barreno, *et al.*, in which they showed for a given upper bound of items classified as false positives it is possible to find the set of combination rules maximizing the number of items classified as true positives[10]. To this end, we are currently designing an analogous experiment using a Bootstrap Percentile Test in precision-recall space[54][55] to determine if our results are equivalent to use of the Neyman-Pearson lemma for proving Boolean AND and OR rules are always part of the optimal set of classifiers.

Our initial test for combining semantic groups across individual systems using a union merge has potential to create more powerful Boolean combination ensembles by further leveraging complementarity between systems. We are currently expanding this strategy with existing Boolean combinations and additional logical operators, such as use of set differences and the NOT operator between multiple semantic groups while excluding others within a particular corpus, or across corpora has exciting potential to truly leverage an ensemble of corpora.

Lastly, use of the NLP-Ensemble-Explorer may be beneficial in terms of being implemented as a step in a pipeline for filtering out spuriously identified concepts from an individual system. Similar methods have been explored by others, including Kim and Riloff, in which they developed a statistical model based on annotations produced by MetaMap to remap to appropriate semantic types within their learning model[46]. Along these lines, we are currently using NLP-Ensemble-Explorer for extraction of UMLS concepts for use in models for covid-19 phenotyping, and for developing word2vec disease models.  Lastly, use of NLP-Ensemble-Explorer for extraction of UMLS concepts for use in creation of weakly supervised data models is also worth exploring.

# CONCLUSION

Use of Boolean ensembling of NLP systems can improve NER and IE performance of individual systems by all measures in general and, differentially, on specific types of named entities. Boolean ensembling affords additional flexibility for applications which seek to maximize either the precision or the recall of NLP, or ideally both. The framework we have developed and presented in this paper can be particularly useful for clinical and translational researchers with a diverse set of information retrieval and extraction needs that can be satisfied by the complementarity and diversity of NLP system ensembles. Our study results also indicate that NER and concept mapping remain to be challenging problems for clinical NLP.

## Acknowledgements

We would like to thank Jen Morgan, Angel Helget, and George Konstantinides for their help with annotating clinical data. We would also like to extend our gratitude to Elizabeth Lindemann for her help in proofreading this manuscript.

## Funding Statement

This research was supported in part by the National Institutes of Health's National Center for Advancing Translational Sciences grant U01TR002062.

## Competing Interests Statement

Dr. Xu and The University of Texas Health Science Center at Houston have research-related financial interests in Melax Technologies, Inc.

## Contributorship Statement

Greg M. Silverman, BS – Lead developer on NLP-Ensemble-Explorer and developer on NLP-ADAPT, study design, data analysis, data interpretation, writing, and critical revision

Raymond L. Finzel, BS – Lead developer on NLP-ADAPT and developer on NLP-Ensemble-Explorer, study design, data analysis, data interpretation, writing, and critical revision

Michael V. Heinz, MD – Developer on NLP-Ensemble-Explorer, study design, data analysis, data interpretation, writing, and critical revision

Jake Vasilakes, MS – Developer on NLP-Ensemble-Explorer, study design, data analysis, data interpretation, writing, and critical revision

Jacob C. Solinsky, BS -- Developer on NLP-Ensemble-Explorer, study design, data analysis, data interpretation, writing, and critical revision

Reed McEwan, MS – Developer on NLP-Ensemble-Explorer and NLP-ADAPT, study design, data interpretation, writing, and critical revision

Benjamin C. Knoll, BS – Developer on NLP-ADAPT and lead developer on BioMedICUS and NLP-TAB, study design, data interpretation, writing, and critical revision

Christopher J. Tignanelli, MD – Study design, data analysis, data interpretation, writing, and critical revision

Hongfang Liu, PhD – Study design, data interpretation, writing, and critical revision

Hua Xu, PhD – Study design, data interpretation, writing, and critical revision

Xiaoqian Jiang, PhD – Developer on NLP-ADAPT, study design, data interpretation, writing, and critical revision

Genevieve B. Melton, MD, PhD – Study design, data interpretation, writing, and critical revision

Serguei Pakhomov, PhD – Study design, data analysis, data interpretation, writing, and critical revision

**Figure Legend**

**Figure 1:** Semantic group hierarchical mapping